\documentclass[preprint,12pt,authoryear]{elsarticle}

\usepackage{amssymb}
\usepackage{amsmath}

\usepackage{textcomp,marvosym}
\usepackage{array}
\usepackage{mathtools}

\journal{Neural Networks}

\begin{document}

\begin{frontmatter}



\title{GASPnet: Global Agreement to Synchronize Phases} 


\author[inst1]{Andrea Alamia \fnref{equal}} 
\author[inst1,inst2]{Sabine Muzellec \fnref{equal}}
\author[inst2,inst3]{Thomas Serre} 
\author[inst1,inst3]{Rufin VanRullen}

\fntext[equal]{These authors contributed equally to this work.}

\affiliation[inst1]{organization={CerCo, CNRS},
            city={Toulouse},
            postcode={31052}, 
            country={France}}
\affiliation[inst2]{organization={Carney Institute for Brain Science, Brown University},
            city={Providence},
            country={USA}}
\affiliation[inst3]{organization={ANITI, Université de Toulouse},
            city={Toulouse},
            postcode={31052}, 
            country={France}}
\begin{abstract}
In recent years, Transformer architectures have revolutionized most fields of artificial intelligence, relying on an attentional mechanism based on the agreement between keys and queries to select and route information in the network. In previous work, we introduced a novel, brain-inspired architecture that leverages a similar implementation to achieve a global “routing by agreement” mechanism. Such a system modulates the network's activity by matching each neuron's key with a single global query, pooled across the entire network. Acting as a global attentional system, this mechanism improves noise robustness over baseline levels but is insufficient for multi-classification tasks, where multiple items are present simultaneously. Here, we improve on this work by proposing a novel mechanism that combines aspects of the Transformer attentional operations with a compelling neuroscience theory, namely, binding by synchrony. This theory proposes that the brain binds together features by synchronizing the temporal activity of neurons encoding those features. This allows the binding of features from the same object while efficiently disentangling those from distinct objects, enabling efficient processing of multiple objects. We drew inspiration from this theory and incorporated angular phases into all layers of a convolutional network. After achieving phase alignment via Kuramoto dynamics, we utilize this approach in an attentional system that enhances operations between neurons with similar phases and suppresses those with opposite phases. We test the benefits of this mechanism on two datasets: one composed of pairs of digits (multi-MNIST) and one composed of a combination of an MNIST item (either a digit or a fashion item) superimposed on a CIFAR-10 image. Our results reveal better accuracy than CNN networks, proving more robust to noise and with better generalization abilities. Overall, we propose a novel mechanism that addresses the visual binding problem in neural networks by leveraging the synergy between neuroscience and machine learning. 
\end{abstract}


\end{frontmatter}


\section*{Introduction}
In recent years, the performance of artificial intelligence systems has improved drastically due to the introduction of the ``Transformer'' architecture. The key innovation is to rely on attentional mechanisms that learn long-distance relations between far-apart elements via key-query vector agreement, such as pixels in an image or words in a sentence. These architectures have had a profound impact on various tasks, ranging from language processing to computer vision \citep{vaswani2017attention, khan2022transformers}. In the latter, Transformer architectures have been proposed as a potential alternative to convolutional neural networks (CNNs) \citep{ramachandran2019stand, mauricio2023comparing, zhao2020exploring}, which have been the \textit{de facto} standard approach for years \citep{kauderer2017quantifying, xu2014scale, krizhevsky2012imagenet, he2016deep}. Recent work has demonstrated that convolutions are a subset of the attention mechanism \citep{cordonnier2019relationship}, and other studies have proposed architectures that successfully combine both approaches \citep{wang2017residual, bello2019attention, vaishnav2022understanding}.

In a previous study, we proposed an architecture named ``GAttANet'' that combines an attentional system with a CNN backbone \citep{vanrullen2021gattanet}. Importantly, the innovative element was inspired by biological brains: on top of a series of convolutional layers, we augmented the network with a \textit{global} attentional system reminiscent of the visual attention processes occurring in the brain (hence the name of the model: GAttANet, for Global Attention Agreement). In the brain, the frontoparietal network (which acts as an attentional system for sensory processes) integrates the information from different sensory areas along the visual hierarchy and sends back modulatory signals according to attentional demand \citep{corbetta2002control, bisley2011neural, itti2001computational}. Similarly, in GAttANet, an iterative attentional mechanism modulates the network's activity by implementing a form of ``routing by agreement''. Similar to the mechanisms in Transformer architectures, each spatial position in the network includes a ``key-query'' vector pair. Though keys are separate for each spatial position, queries are pooled together across the entire network in a single, global query. Attentional modulation is then computed by matching each neuron's key with the global query. In our previous study, we showed that this approach is beneficial for improving accuracy over an equivalent baseline \citep{vanrullen2021gattanet} in classification tasks.
However, the ``routing by agreement'' mechanism we proposed in GAttANet is optimized for classification tasks in which there is one object at a time. In these cases, the global query provides a unique hypothesis for such an item, while the keys serve as candidate features from different parts of the network. Similar to an attentional mechanism, the keys that match the hypothesis represented by the global query are enhanced, whereas the others are decreased.  However, in more ecological conditions, several objects may be present at once, potentially confounding the global agreement mechanism. In these cases, the global query would select, at best, only one of the several items while ignoring the others, or, in the worst case, it would produce erroneous hypotheses by incorrectly binding together features from different objects. In human attention, combining features from different objects into a single, false perception is known as the illusory conjunction problem \citep{treisman1982illusory}. Interestingly, it has been proposed that the brain employs different mechanisms to overcome these attentional confusions when multiple objects are present. One such mechanism is the ``binding-by-synchrony'' hypothesis, which proposes that the brain solves this problem by synchronizing the temporal activity of neurons that encode features from the same object \citep{singer2007binding, singer1999neuronal, treisman1996binding}, thus disentangling different objects' features via distinct temporal patterns. 

Here, we took inspiration from this theory and previous work in neuroscience \citep{hummel1992dynamic, roskies1999binding, roelfsema2023solving, garrett2024binding} and in machine learning \citep{lowe2024binding, reichert2013neuronal, gopalakrishnan2024recurrent, stanic2023contrastive, muzellec2025enhancing} to implement a process similar to binding-by-synchrony. Specifically, we introduced a phase value for each spatial location in the network (in the convolutional layers) and for each node in the dense layers. We then leveraged the attentional mechanism to modulate the network's activity via phase synchronization. Such synchronization then modulates each neuron's activation amplitude: phase synchrony determines whether the modulation of each neuron's activity is positive (in phase) or negative (out of phase) relative to the activity of all other neurons in the network. We tested our proposed architecture on two datasets, one with two digits on the image (multi-MNIST)~\citep{lecun1998gradient, sabour2017dynamic} and one with a digit or fashion item~\citep{xiao2017_online} superimposed on a CIFAR-10~\citep{krizhevsky2009learning} background. Our results reveal that phase synchronization through the attentional mechanism and global agreement could be a helpful mechanism for binding features from distinct objects, leading to improved classification when multiple items are present, and to better accuracy and generalization abilities in artificial networks. 

\section*{Materials and methods}
\subsection*{Datasets}
We evaluate our approach using two custom datasets to assess classification performance under varying levels of noise and scale distortion.
\subsubsection*{Multi-MNIST Dataset}
The first dataset is a variant of the Multi-MNIST dataset, inspired by~\citep{sabour2017dynamic}. Each image consists of two non-overlapping MNIST digits~\citep{lecun1998gradient}, ensuring that the same class of digit never appears more than once in an image. The images are grayscale and measure $32\times 32$ pixels. Each image is accompanied by a segmentation mask that matches the image dimensions and specifies whether each pixel corresponds to the background, the first digit, or the second digit.
The classification task requires predicting both digits using a two-hot encoding scheme. Additionally, the masks are used to optimize a phase segmentation objective (similar to ~\citep{muzellec2025enhancing}) at the first layer. Training images are noise-free, while test images include additive Gaussian or salt-and-pepper noise at varying intensities. Specifically, Gaussian noise is applied with standard deviations of $0.15$, $0.25$, $0.35$, $0.45$ and $0.6$ and salt-and-pepper noise is introduced at corruption levels of $0.01$, $0.03$, $0.06$, $0.1$ and $0.2$. The masks are not used during test.
\subsubsection*{CIFAR-MNIST and CIFAR-FashionMNIST Datasets}
The second and third datasets incorporate images from the CIFAR~\citep{krizhevsky2009learning} dataset as backgrounds, with either an MNIST digit (CIFAR-MNIST) or a FashionMNIST item (CIFAR-FashionMNIST) overlaid in transparency. This composition ensures that both the background and the overlaid object remain visible. Corresponding segmentation masks are generated to label each pixel as belonging to either the background or the overlaid object. In this setting, the phases are expected to learn foreground-background segmentation.
The classification task requires jointly identifying the CIFAR object and the overlaid MNIST or FashionMNIST item. During training, the overlaid item is consistently $28\times28$ pixels, located at various positions within the $32\times 32$ CIFAR image. In the test set, this item undergoes progressive downscaling to assess the model's robustness to size variations, with test sizes of $24\times 24$, $20\times 20$, $17\times 17$, and $14 \times 14$ pixels.

These datasets enable us to systematically evaluate classification performance under varying levels of noise and object size changes, providing insight into the models' generalization capabilities.
\subsection*{The model}

\subsubsection*{Architecture}
\begin{figure*}[hbt!]
    \centering \includegraphics[width=1\textwidth,keepaspectratio=true]{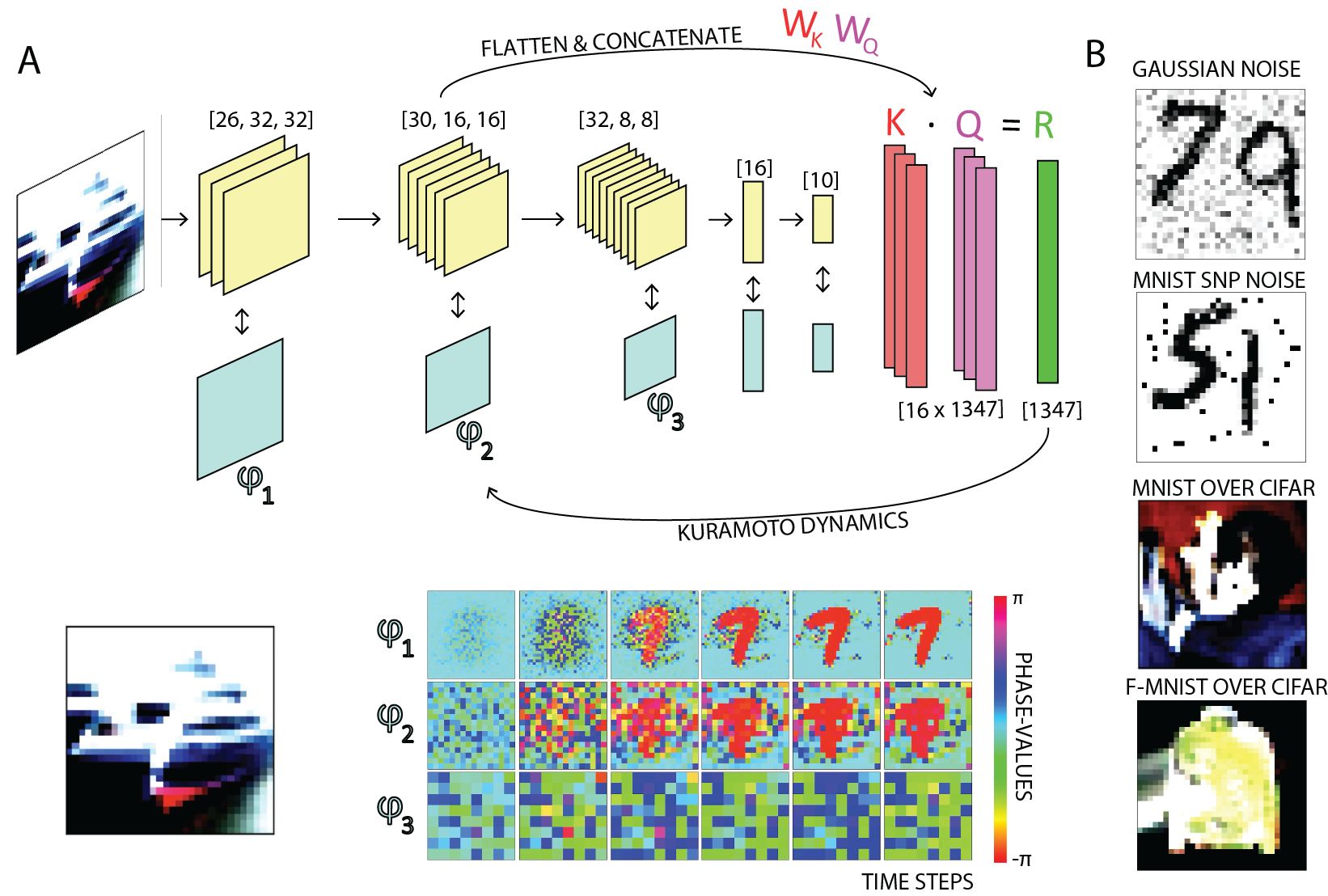}
    \caption{{\bf Proposed architecture and image examples.} A) GASPnet is an augmented 3-layer convolutional network with phases and the key-query attentional system. Each layer's activity is projected into the key-query space, where the key-query dot product provides the couplings (R, shown in green in the figure) that influence phase synchronization over time steps via Kuramoto dynamics. The lower panel shows how phases synchronize over time steps. The over-imposed digit '7' is synchronized and in anti-phase with the rest of the figure. B) Example images from the dataset we used to test the network. First, two panels show two-digit MNIST with Gaussian and Salt and Pepper noise. The last two panels show an MNIST image, either a digit or a fashion item, over-imposed on a CIFAR-10 image.}
    \label{fig:01}
\end{figure*}

GASPnet (Global Agreement to Synchronize Phases) is based on a feedforward architecture composed of three convolutional layers and two dense layers, as shown in Fig~\ref{fig:01}. 

The backbone consists of three sequential $3 \times 3$ convolutional layers, each followed by ReLU activation, 2D MaxPooling, and instance normalization~\citep{ulyanov2016instance}.
These layers have $26$, $30$, and $32$ channels, respectively. The extracted convolutional features are then flattened and passed through two fully connected layers with $16$ and $10$ units. For the CIFAR-MNIST and CIFAR-FashionMNIST datasets, the model includes two separate classification heads (final dense layers), one for each dataset.

\subsubsection*{Phase modulation}
Different from a usual feedforward network, we introduce a phase $\phi_{i}$ shared across neurons with the same spatial position in each layer $i$ (light blue layers in Figure~\ref{fig:01}). Importantly, the phases play a role in modulating the result of the forward pass: whether they are in phase or anti-phase, i.e., their difference is equal to 0 or $\pi$, they enhance or reduce the network activity, respectively. Specifically, the cosine of the difference of subsequent phases $\phi_{i}$ multiplies the results of the convolution, as described in Eq~\ref{eq:convPhase}.

\begin{eqnarray}
\label{eq:convPhase}
  m_{i+1}(x) &= \int{[1+ \alpha_{F} cos(\phi_{i+1}(x)-\phi_{i}(x-t))]*m_{i}(x-t)*W_c(t)dt}
\end{eqnarray}

where $W_c(t)$ represents the learned convolutional weights, and $\alpha_{F} \in [0,1]$ is a hyperparameter controlling the influence of phase modulation. A value of $\alpha_{F} = 1$ ensures that only spatial regions with synchronized phases propagate to the next layer, whereas $\alpha_{F} = 0$ reduces the model to a standard feedforward convolutional network by nullifying phase effects.

Phase modulation is applied at each convolutional block -- after ReLU activation, MaxPooling, and normalization -- by incorporating upsampled phase values from the current layer and interacting with those of the previous layer.

Interestingly, such phase modulation can be directly applied to the convolution using its usual operator and trigonometric transformations, as shown in Supplementary Appendix \ref{S1_Appendix}. A similar equation (Eq~\ref{eq:densePhase}) describes the effect of the phase synchronization on the dense layers:

\begin{eqnarray}
\label{eq:densePhase}
  m_{i+1}(x) &= [1+ \alpha_{F} cos(\phi_{i+1}-\phi_{i})]*(m_{i}(x) * W_i)
\end{eqnarray}
where $W_i$ represents the learned weights of the fully connected layer $i$.

\subsubsection*{Phase synchronization}
The phases modulate the network's activity, functioning as an attentional mechanism that enhances or suppresses the relevance of specific input regions during sensory processing. This modulation is governed by their degree of synchrony, following a variation of the Kuramoto model \citep{kuramoto1975self}.

\begin{eqnarray}
\label{eq:updatePhase}
  \phi_{i}(t+1) &= \phi_{i}(t) + \lambda \frac{\sum_{j=1}^{N}{r_{ij}sin(\phi_{j}(t)-\phi_{i}(t))}}{\sum{|r_{ij}|+\epsilon_\theta}},
\end{eqnarray}
where $\lambda \in \mathbb{R}$ is a hyperparameter modulating the phase update at each timestep and $N$ is the total number of phases within the network. According to Eq~\ref{eq:updatePhase}, each phase is updated over time, based on its neighbors' states, and weighted by a coupling term $r$.

Specifically, the couplings $r_{ij}$ are defined as the dot product between the activities $m_{i}$ projected in a Key-Query space (denoted Q and K in Figure~\ref{fig:01}).
The couplings of convolutional layers are modulated by a neighborhood matrix, which enhances the connection strength between spatially close phases by a factor of $\tau$ and reduces the connection strength between spatially distant phases by a value $\epsilon$. Finally, to compensate for the significantly smaller number of neurons in the dense layers compared to the convolutional layers, the contribution of the dense layers to the coupling is amplified by a factor $\kappa$.
\begin{eqnarray}
\label{eq:coupling}
  r_{ij} = (<W_{q}*m_{i},W_{k}*m_{j}>*N_{i,j}-\epsilon)/\kappa
\end{eqnarray}
Where, 
\begin{equation}
N_{ij} = \begin{cases*}
  \tau, & if $i = j \pm 1$ and $j = i \pm 1$ and $l_i = l_j$,\\ 
  1,                    & otherwise.
\end{cases*}
\end{equation}
And, 
\begin{equation}
\kappa = \begin{cases*}
  100, & if $i,j \in l_{d_1} or l_{d_2}$ dense layers,\\ 
  1,                    & otherwise.
\end{cases*}
\end{equation}
With $l_i$ denoting the layer of neuron $i$ and $l_{d_n}$ denoting the dense layer $n$ with $n \in {1,2}$.

Overall, for a single image presentation, the phase dynamics involve projecting the forward activity into the Key-Query space, computing the coupling matrix between phases, and modulating it based on the spatial organization of each layer and the model's depth. The resulting phases then influence network activity in subsequent timesteps, and this process is iterated over $T$ timesteps.

\subsection*{Baseline}
To evaluate the contribution of phase modulation to GASPnet's performance, we compare it with a baseline model that has a matching number of parameters. This baseline is a standard feedforward network with a similar architecture (yellow layers in Figure~\ref{fig:01}). To account for the additional parameters introduced by the Key-Query space projections, we increase the number of convolutional channels in the baseline model to $28$, $32$, and $35$, respectively.

\subsection*{Training}
During training and testing, GASPnet's activity is initialized with a standard feedforward pass, where phase modulation is not applied. During this step, the network's activations are projected into the Key-Query space, while the phases are independently initialized from a standard normal distribution (mean is $0$ and variance is $1$ radians). Optionally, the phase initialization can be learned as a pixel-wise trainable parameter, controlled by a hyperparameter. 

In subsequent timesteps, the phases begin to modulate the network's activity, and all parameters -- including the layer weights and the Key-Query projections -- are jointly optimized through gradient descent.

Both models are trained to minimize a Cross-Entropy loss for multi-object classification, alongside a synchrony loss for GASPnet (Eq.~\ref{eq:synch_loss}) that encourages phase synchronization in the first layer before propagating to subsequent layers (similar to~\citep{muzellec2025enhancing}).

\begin{equation}\label{eq:synch_loss} synLoss(\phi) = \frac{1}{2} \left( \frac{1}{G} \sum_{l=1}^{G} V_l(\phi) + \frac{1}{2G} \Bigl| \sum_{l=1}^{G} e^{i\langle\phi\rangle_l} \Bigr|^2 \right) \end{equation}

where $V(\phi)$represents the circular variance, $\langle\phi\rangle$ denotes the average phase within a group, and $G$ is the number of groups. The groups are defined using the ground-truth masks described in the dataset section.  The first term of the loss function measures intra-cluster synchrony, ensuring that the phases within each group converge to the same value. The second term enforces inter-cluster desynchrony, ensuring that the centroids of different phase clusters on the unit circle remain distant and ideally cancel each other out.

The contribution of the synchrony loss to the overall optimization is controlled by a hyperparameter $\omega$, which modulates its relative importance in the loss function.

Both GASPnet and its baseline are trained for $25$ epochs with batches of size $32$, using Adam optimizer~\citep{kingma2014adam}, a learning rate of $0.0005$, and weight decay of $0.00001$. Both classification loss and $synLoss$ are computed and combined at the last timestep. All experiments are implemented in Pytorch 1.13 \citep{paszke2017automatic} and run on a single NVIDIA TITAN Xp.

The values of the hyperparameters are determined using a hyperparameter search and chosen using a held-out validation set. They are reported in Table~\ref{tab:hyperparams}.

\begin{table}[h]
    \centering
    \begin{tabular}{l|ccccccccc}
        \hline
        Dataset & $\alpha$ & $D$ & $\lambda$ & $\kappa$ & $\epsilon$ & $\tau$ & $\omega$ & Init & $T$ \\
        \hline
        Multi-MNIST & 1 & 32 & 1 & 100 & -0.9 & 5 & 0.5 & FALSE & 6 \\
        CIFAR10-FashionMNIST & 0.8 & 16 & 4 & 100 & -0.7 & 1 & 10 & TRUE & 6 \\
        CIFAR10-MNIST & 0.8 & 16 & 4 & 100 & -0.7 & 1 & 10 & TRUE & 6 \\
        \hline
    \end{tabular}
    \caption{Hyperparameters used for each dataset. $\alpha$ is the phase influence on the activity. $D$ is the dimension of the Key-Query space. $\lambda$ modulates the phase updates per timestep. $\kappa$ is a multiplicative factor of the contribution of the dense layers in the coupling matrix. $\epsilon$ impacts the coupling strength of spatially distant phases while $\tau$ modulates the coupling strength of nearby phases. ``Init'' determines whether the phase initialization is learned, and $T$ represents the number of timesteps (not optimized as a hyperparameter).}
    \label{tab:hyperparams}
\end{table}
We additionally conduct an ablation study to assess the importance of $\alpha$, $\omega$, and the location-based parameters ($\kappa$ and $\epsilon$). These hyperparameters were selected for ablation as they were explicitly constrained to influence the model during hyperparameter search, unlike the others, which included the option of being deactivated (set to $0$, $1$, or False). The only exception is $D$, for which we explored values from $8$ to $32$, in increments of $8$.

\subsection*{Statistical analysis}
After selecting the optimal set of hyperparameters, we train both GASPnet and its baseline using ten different weight initializations in the multi-MNIST dataset. We evaluate performance by comparing the accuracies of GASPnets with those of the baseline via t-tests. Regarding the experiments with the CIFAR-MNIST and CIFAR-FashionMNIST datasets, we first obtained the best model on the validation set from five different weight initializations of GASPnet. We then tested for each time step whether the GASPnet accuracy was significantly different than those obtained by ten different weight initializations of the baseline. All p-values were corrected for multiple comparisons using the Benjamini and Yekutieli procedure for false discovery rate (FDR) \citep{benjamini2001control}. All statistical analyses were performed in MATLAB 2018, using standard built-in functions.

\section*{Results}
We assessed the advantages of phase synchronization via a global agreement in two distinct conditions. First, we evaluated the changes in accuracy over time steps in the multi-MNIST dataset on clean images, as well as with noise corruptions (i.e., additive Gaussian noise and Salt and Pepper; see \ref{fig:01}B for some examples). Next, we investigated whether phases can be instrumental in disentangling two over-imposed images from distinct datasets. Specifically, we quantified the accuracy in classifying MNIST and fashion-MNIST items superimposed on CIFAR-10 images. In this case, we used a two-head architecture to classify both MNIST and CIFAR-10 images (see methods for details).  

\subsection*{Robustness to noise corruption}
\begin{figure*}[hbt!]
    \centering \includegraphics[width=1\textwidth,keepaspectratio=true]{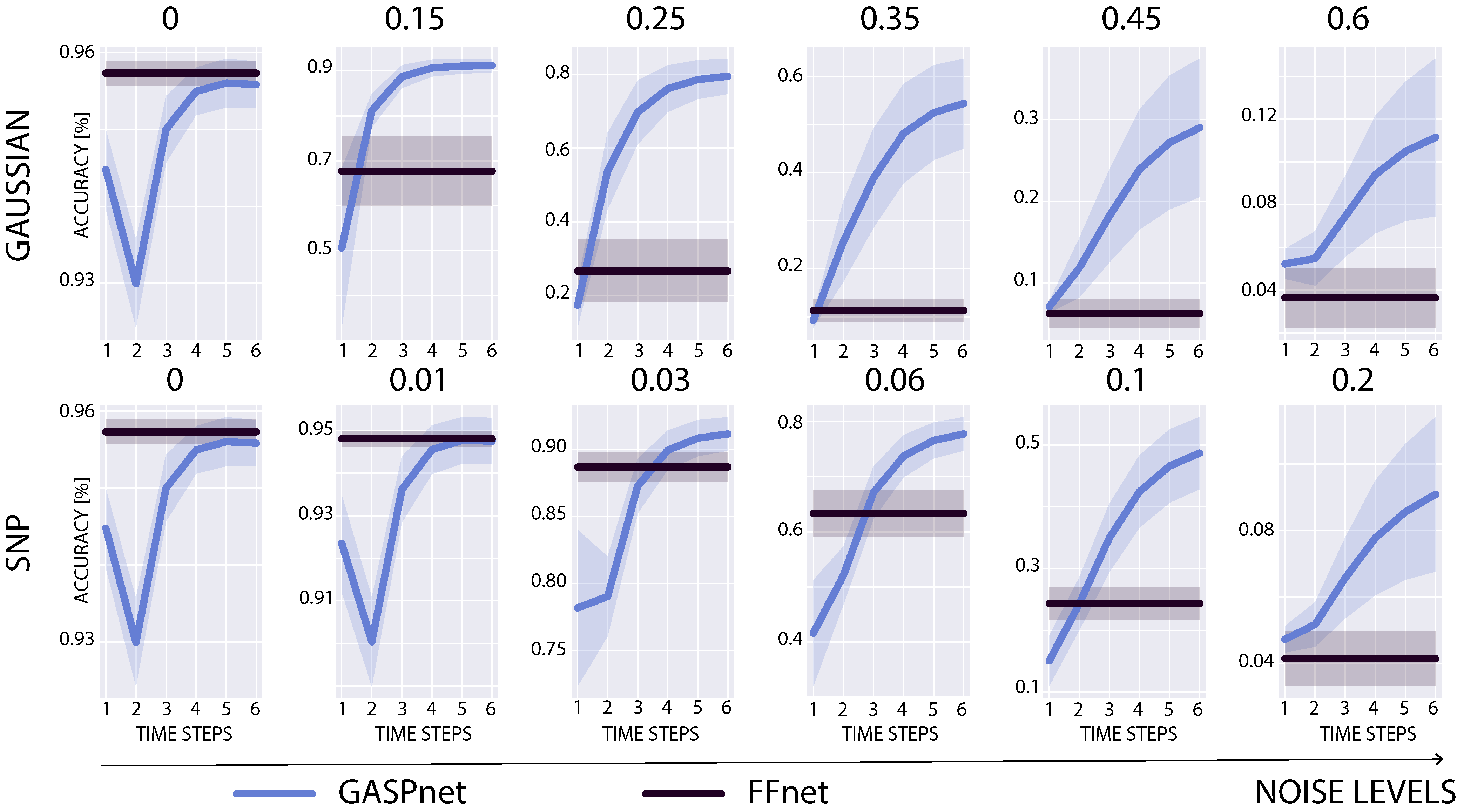}
    \caption{{\bf Robustness to noise.} Accuracy over time steps for GASPnet (in blue) and forward networks having an equivalent number of parameters. The upper and lower plots show the results for Gaussian and Salt and Pepper (SNP) noise, respectively. GASPnet performs similarly for low levels of noise (left panels), and significantly better for higher levels of noise (right panels). The noise level is specified on top of each panel.}
    \label{fig:02}
\end{figure*}

We first compared GASPnet's accuracy to that of a forward network with an equivalent number of parameters on clean images. We considered the multi-MNIST dataset, composed of two non-overlapping MNIST digits randomly placed in a $32\times 32$ image (see figure \ref{fig:01} for some examples, and the methods section for details). As shown in Figure \ref{fig:02}, GASPnet performs similarly to its equivalent forward network on clean images, with close to 100\% accuracy, and increases over time steps after a slight drop. Interestingly, as the noise level increases, the contribution of the phases becomes more critical. Over time steps, the accuracy of GASPnet regularly improves, outperforming the forward network after a few time steps. Note that neither GASPnet nor the forward network baseline was exposed to noise corruption during training; the robustness of GASPnet is thus an emergent property. These results were supported by a series of t-tests corrected for multiple comparisons, in which we compared GASPnet accuracy at each time step with the baseline: regarding Gaussian noise, for all noise levels, we found a significant difference between accuracies from the third time step (all FDR corrected $p<0.05$, with larger significance at later time steps). Regarding the Salt and Pepper noise, we obtained similar results: the difference between GASPnet and the baseline is statistically significant for the last three noise levels in the last three time steps (all FDR-corrected $p < 0.05$). Importantly, we found significant results despite the relatively limited statistical power (i.e., ten samples at each time step). If, on the one hand, this limitation is due to our explicit choice to reduce the number of simulations for energy and computational costs, on the other, it corroborates the reliability of our results. 

\subsection*{Phases disentangle MNIST item over-imposed on CIFAR-10 images}

\begin{figure*}[hbt!]
    \centering \includegraphics[width=0.4\textwidth,keepaspectratio=true]{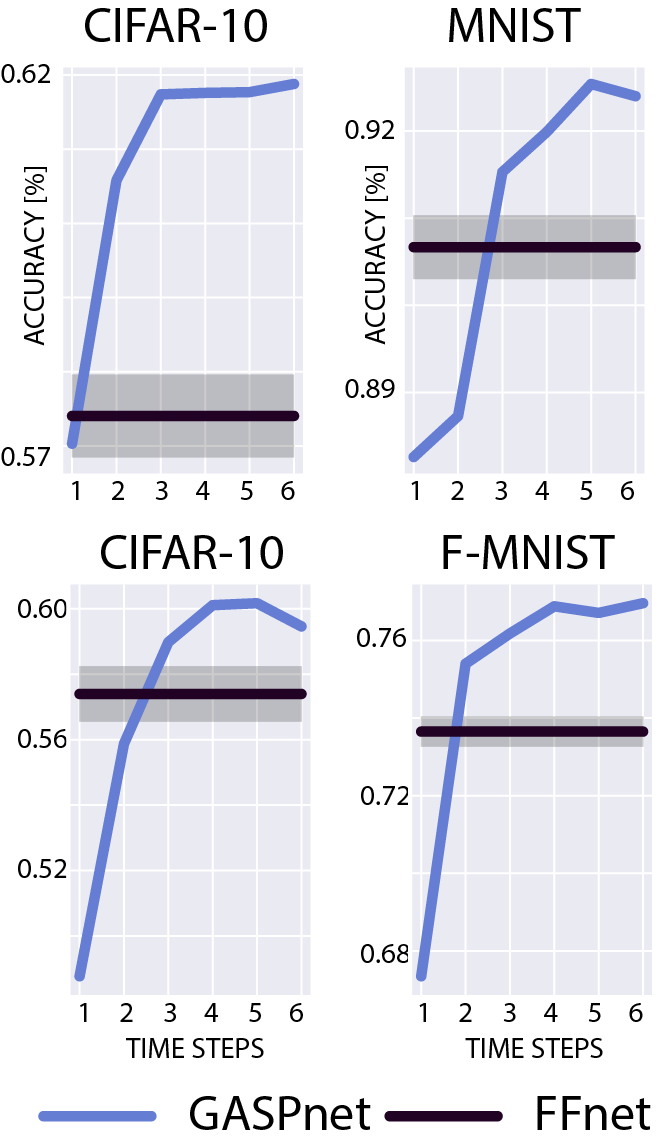}
    \caption{{\bf Multi heads results on CIFAR and MNIST.} Results for GASPnet (in blue) and an equivalent forward network (in black) as accuracy over time steps for the MNIST over CIFAR-10 images. GASPnet outperforms the forward network on both classification tasks.}
    \label{fig:03}
\end{figure*}

We then investigated the contribution of the phases when an MNIST item was superimposed on a CIFAR-10 image. In this dataset, we tested an architecture with two heads, thus allowing the classification of both the MNIST item (either a digit or a fashion item) and the CIFAR-10 image. Figure~\ref{fig:03} shows that GASPnet performs slightly better than the equivalent forward network in both datasets and for both classification tasks. As expected, the accuracy increases over the six timesteps, outperforming the baseline at the last step. The statistical analysis corroborates these results: regarding the digit MNIST dataset, we found a significant difference between architectures in classifying CIFAR-10 images ($p<0.05$ from the third time step) but only a trend in digit classification ($p=0.1$ for the first and last two time steps). Regarding the fashion MNIST dataset, we found a significant difference in classifying fashion items after the third time step (all $p<0.05$) but no difference in the classification of CIFAR-10 images (all $p>0.1$). All in all, these results confirm that the phases are beneficial in increasing the accuracy of the network, either for the CIFAR or the MNIST datasets, and allow for reliably disentangling the two images, as shown in Figure~\ref {fig:01}A. In the following, we investigate the network's ability to generalize to over-imposed items with different sizes.

\subsection*{Superimposed images: generalizing to different sizes}

\begin{figure*}[hbt!]
    \centering \includegraphics[width=0.8\textwidth,keepaspectratio=true]{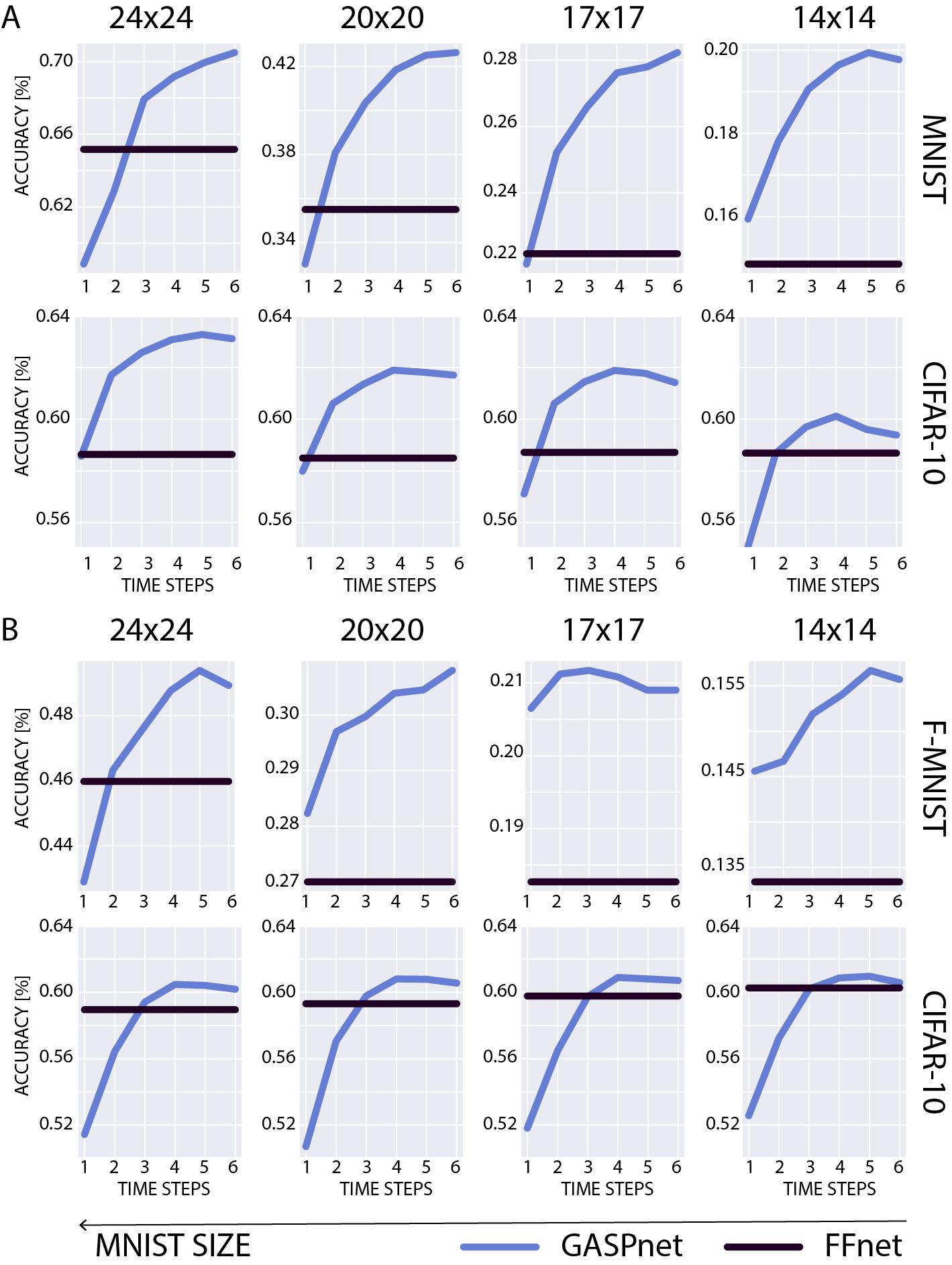}
    \caption{{\bf Results varying the size of the over-imposed MNIST item.}  Generalization results for GASPnet (in blue) and the forward network (in black) when varying the size of the MNIST item over-imposed on the CIFAR-10 image. Both networks were trained on a dataset composed of $28\times 28$ MNIST over-imposed on a CIFAR-10 image. GASPnet (in blue) systematically outperforms the forward network (in black) in both classification tasks (MNIST and CIFAR-10, upper and lower rows, respectively). Panels in A show the results for MNIST, panels in B for Fashion-MNIST}
    \label{fig:04}
\end{figure*}

We then investigated whether a network trained with a superimposed image of a given dimension could generalize to images of a different size. Specifically, we first trained GASPnet and the equivalent baseline forward networks on $32\times 32$ CIFAR-10 images with a $28\times 28$ superimposed MNIST item (either a digit or a fashion item). Next, we assessed the accuracy of both architectures in classifying both images but with smaller MNIST items than during the training set (i.e., $24\times 24$, $20\times 20$, $17\times 17$, and $14\times 14$). As shown in Figure~\ref{fig:04}, GASPnet consistently outperforms the corresponding baseline in classifying both MNIST items and CIFAR-10 images, demonstrating that it generalizes across object sizes without compromising performance on CIFAR background classification. Statistical analyses corroborate these results: in both datasets, we found a significant difference between architectures in classifying MNIST items after the third time step ($p<0.01$). This increase in performance was not achieved at the expense of classification accuracy on CIFAR-10 images, which remained at comparable levels for both architectures.

\begin{figure*}[hbt!]
    \centering \includegraphics[width=.65\textwidth]{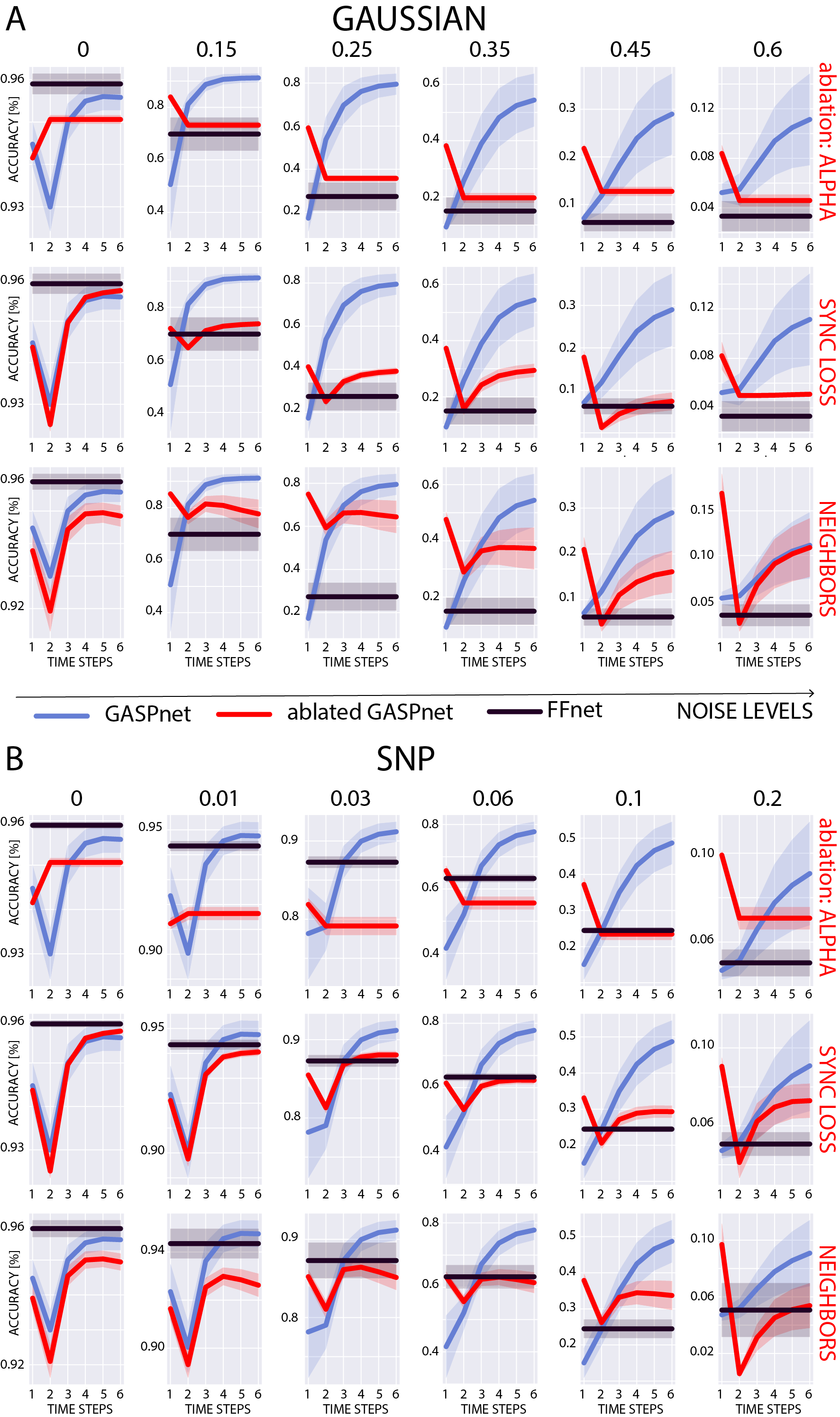}
    \caption{{\bf Ablation Results.} Accuracy over time steps for different ablations in GASPnet (in red), the network without ablations (in blue), and forward networks having an equivalent number of parameters. The plots in A and B show the results for Gaussian and Salt and Pepper (SNP) noise, respectively. GASPnet performs similarly for low levels of noise (left panels), and significantly better for higher levels of noise (right panels). The noise level is specified on top of each panel.}
    \label{fig:05}
\end{figure*}

\subsection*{Ablation studies}
To understand the individual contributions of key components in GASPnet, we performed ablation studies focusing on phase modulation strength ($\alpha$), $synLoss$ weight ($\omega$), and spatial neighborhood coupling ($\kappa$ and $\epsilon$). These experiments were conducted on the Multi-MNIST dataset under both Gaussian (Fig.~\ref{fig:05}A) and Salt-and-Pepper noise conditions (Fig.~\ref{fig:05}B).

Across all conditions, removing any of these components consistently resulted in a noticeable decrease in performance compared to the full model. In particular, the gap between the complete model and its ablated counterpart widened over time, especially at higher noise levels. This trend was observed for the phase modulation strength ablation, the synchrony loss ablation, and the spatial neighborhood coupling ablation. These results were supported by statistical analyses comparing the full model and the different ablated ones. Specifically, the analyses confirmed a significant difference (corrected $p<0.05$) in the case of phase modulation ablation from the third noise level (for Gaussian noise) after the third timestep, and after the second time step for the second noise level (for salt-and-pepper noise). We also found a significant difference when the synchrony loss was removed, both for Gaussian noise from the second time steps and the second noise level ($p < 0.05$), and for Salt-and-Pepper noise at noise levels of 0.06 and 0.1 from the fourth time step. Lastly, we found a significant difference between the full model and the one without the spatial neighborhood coupling in both noise types from the third noise level (but not the last one) and from the fourth time step. These findings highlight that all three components contribute to the progressive improvement of the model’s accuracy over time.

Overall, the ablation results confirm that the synchrony dynamics and spatial coupling mechanisms are integral to GASPnet’s robustness. Removing any of these mechanisms impairs the network’s ability to refine its representations across timesteps, especially under challenging, noisy conditions.

\section*{Discussion}

In this work, we propose a novel, brain-inspired mechanism that combines phase synchronization and global attention, drawing inspiration from previous neuroscience research on visual attention \citep{corbetta2002control, itti2001computational} and neural synchrony \citep{singer1999neuronal, singer2007binding}. Specifically, our model integrates phase dynamics into a convolutional neural network and leverages the global attentional mechanism to synchronize their activity, modulating the network's operations. In two experiments with different datasets, we demonstrated that such an architecture is more robust to Gaussian and salt-and-pepper noise corruption, as well as better at generalization than its equivalent forward network. 

Our architectural choices are grounded in the neuroscience hypothesis of binding-by-synchrony \citep{singer2007binding, singer1999neuronal, treisman1996binding}. In particular, we equipped a convolutional network with a binding mechanism based on phase synchronization, which is achieved via Kuramoto dynamics \citep{kuramoto1975self}. In this light, the phases represent object assignment: for each position in the convolutional layers and each feature in the dense ones, the phases are assigned to distinct items in the image. Importantly, as in the GAttANet architecture, the attentional system operates globally across the whole network. Crucially, the phase agreement is computed across all layers to bind hierarchically distinct features into a single object, thus achieving a binding that ranges from low-level spatial properties to high-level category-related features. 
We hypothesized that such a mechanism would be helpful in solving the illusory conjunction problem \citep{treisman1982illusory}, which occurs when multiple items are present concurrently and the system erroneously binds features from different objects into a single representation, thus generating incorrect percepts. Our results show that phase synchronization, besides improving against noise robustness, benefits the classification with multiple objects, successfully addressing the illusory conjunction issue.

The idea of introducing phases in machine learning is properly outlined in the mathematical framework of complex-valued neural networks. Although complex-valued models are a relatively new and emerging field in machine learning, previous works have investigated these architectures in various tasks, generally not from a biological perspective \citep{lee2022complex, bassey2021survey}. Trabelsi and colleagues \citep{trabelsi2017deep} adapted most commonly used operations from the real to the complex domain, including activation functions and normalization. Other studies compared classical implementations with complex-valued ones, investigating how different choices influence the network's accuracy and behavior \citep{guberman2016complex, monning2018evaluation, yadav2023fccns}. Other studies implemented complex-valued networks or mechanisms aimed at synchronizing the model's activity, similar to a binding mechanism. Early work focused on simple datasets to solve visual segmentation tasks via phase synchronization \citep{zemel1995lending, rao2008unsupervised, weber2005image}, exploring how synchronization can be achieved through horizontal or top-down connections, and in combination with sparsity constraints \citep{rao2011effects, ravishankar2010objective}. However intriguing, previous work has mostly focused on datasets that present only one item per image, thereby synchronizing the phases to separate the foreground from the background.  On the contrary, a study from Reichert \& Serre \citep{reichert2013neuronal} leveraged phase synchrony to disentangle different and overlapping objects in an image in Boltzmann machines, proving that phase synchrony can be beneficial in more ecological situations. Similarly, a more recent study from Lowe and colleagues \citep{lowe2022complex} introduced phase synchrony in an autoencoder architecture. The authors showed that the model, trained for reconstructing multi-object images, leverages phase synchrony to reconstruct the image, outperforming its real-value equivalent. Such an autoencoder architecture was also tested on datasets with more objects and in combination with a contrastive loss for the phase synchrony \citep{stanic2023contrastive}, as well as with complex weight and recurrent activations \citep{gopalakrishnan2024recurrent}. 

In previous work, using a different architecture, we demonstrated that phase synchrony can benefit multi-object classification under overlapping and noisy conditions \citep{muzellec2025enhancing}. In that work, we investigated how phase synchrony can be induced in a complex-valued convolutional neural network and assessed the role of feedback during synchronization. In GASPnet, we further generalized these results by introducing a novel element: the global attentional system, which leverages phase synchronization across the entire network to modulate its activity. 
Finally, synchrony has been successfully applied to other visual tasks beyond classification. Such work includes object tracking using a complex-valued recurrent neural network that tracks the changing features of a given target over time \citep{muzellec2024tracking}, as well as applications in adversarial robustness, uncertainty estimation, and logical reasoning tasks \citep{miyato2024artificial}. Other works, using similar dynamics based on Kuramoto oscillators, have been successfully used in image segmentation tasks \citep{liboni2025image} as well as simple video predictions \citep{benigno2023waves}.

From a mathematical perspective, GASPnet is not equivalent to a complex-valued neural network because each neuron in our network is not represented by a magnitude and a phase (or, equivalently, by a real and imaginary component). Instead, GASPnet has a real activation/magnitude for each neuron and a phase value that is shared between several neurons at the same position in a convolutional layer. From a biological perspective, one can speculate that such a group could roughly correspond to a cortical column, at least in retinotopic regions or spatially organized convolutional layers. Nonetheless, our model finds its place in the growing literature on complex-valued neural networks and stands out as it proposes a brain-inspired global attentional mechanism to synchronize phases. Moreover, our proposed model combines the attentional mechanisms of Transformer architectures with phase synchronization, thereby paving the way for future work that aims to implement Transformers in the Complex domain.

Our model can be further improved in several directions in the future. In this study, we presented the first proof-of-concept model that combines phase synchronization and global attention, and we tested it on relatively small datasets. A first, obvious way to expand our work is to train GASPnet on larger datasets, for longer time steps, and with a deeper model. This would corroborate the benefits of phase synchronization when performing object classification with several objects and in noisy conditions. Additionally, assessing the model's performance in other tasks, such as visual reasoning, would be a promising direction. Interestingly, previous experimental work demonstrated the role of oscillations and phase synchrony in several cognitive functions \citep{pockett2009eeg, palva2010neuronal}, including visual reasoning \citep{alamia2021differential, buschman2007top, benchenane2011oscillations}. 

Furthermore, our model could be further integrated with other frameworks from the cognitive sciences. For example, it would be possible to introduce Predictive coding (PC) dynamics to the current architecture. PC is a well-explored framework in neuroscience \citep{rao1999predictive, shipp2016neural} and machine learning \citep{choksi2021predify, lotter2016deep}, in which a given layer reconstructs (i.e., predicts) the activity of the hierarchically lower layer, and it has been shown to improve noise robustness and generalization \citep{alamia2023role, choksi2021predify}.
Although GASPnet does not implement any predictive or recurrent dynamics, it would be interesting to leverage phase synchrony to achieve this reconstruction objective. In this approach, the reconstruction objective drives phase synchronization, further promoting the disentanglement of distinct objects' features, in an approach similar, in principle, to the autoencoder proposed by Lowe and colleagues \citep{lowe2022complex}. Additionally, it would be possible to implement an endogenous, top-down attentional mechanism, thereby manipulating visual expectations within the network by setting the system's queries to a predetermined value. This approach would encode prior assumptions and represent the endogenous focus of attention, being beneficial in a visual search task. 
Additionally, it is possible to expand our architecture to include a multimodal approach, in which multiple modalities are integrated simultaneously. For example, one can imagine several convolutional networks processing different sensory inputs while being modulated by the same key-query space in which all the sensory information from all networks is embedded. This approach would effectively implement a multi-sensory attentional system, possibly similar to brain mechanisms \citep{talsma2010multifaceted}.

\section*{Conclusion}
In this work, we introduce a new architecture that combines phase synchronization with an attentional system inspired by the Transformer architecture. We demonstrated that this model achieves better performance and generalization than its equivalent forward network in classification tasks, especially in noisy conditions. Altogether, our results support the use of brain-inspired approaches to enhance current AI systems.

\paragraph*{S1 Appendix.}
\label{S1_Appendix}
{\bf Convolution including phase contribution.} Our objective is to apply the cosine of the phase difference for every pair of neurons affected by the convolution operator (with kernel $K$):

\begin{align*}
  Layer_2(x) &= \int{[1+cos(\varphi_2(x)-\varphi_1(x-t))]*Layer_1(x-t)*K(t)dt}\\
\end{align*}

This is challenging because it seems we should create a new convolution layer/operation. But in fact, we implemented it in GASPnet with ``standard" convolution and trigonometry (with complex phase values), as shown below.

Let's call 
\begin{align*}
C &=\int{Layer_1(x-t)*K(t)dt}\\
C_{cos} &=\int{cos(\varphi_1(x-t))*Layer_1(x-t)*K(t)dt}\\
C_{sin} &=\int{sin(\varphi_1(x-t))*Layer_1(x-t)*K(t)dt}
\end{align*}

Each can be computed by first multiplying $Layer_1$ activity by the corresponding $cos$ or $sin$ term, then applying standard convolution. Then, expressing the cosine as the real-part of a complex number, and taking the real-part out of the integral, we get:

\begin{align*}
  Layer_2(x) &= \int{[1+Re(e^{-i(\varphi_2(x)-\varphi_1(x-t))})]*Layer_1(x-t)*K(t)dt}\\
  &= \int{[1+Re(e^{i\varphi_1(x-t)-i\varphi_2(x)})]*Layer_1(x-t)*K(t)dt}\\
  &= \int{Layer_1(x-t)*K(t)dt} + Re(\int{e^{i\varphi_1(x-t)-i\varphi_2(x)}*Layer_1(x-t)*K(t)dt})\\
  &= C + Re(e^{-i\varphi_2(x)}*\int{e^{i\varphi_1(x-t)}*Layer_1(x-t)*K(t)dt})\\
  &= C + Re(e^{-i\varphi_2(x)}*\int{[cos(\varphi_1(x-t))-i*sin(\varphi_1(x-t))]*Layer_1(x-t)*K(t)dt})\\[2pt]
  &= C + Re(e^{-i\varphi_2(x)}*(C_{cos}-i*C_{sin}))\\[8pt]
  &= C + Re[(cos\varphi_2(x)+i*sin\varphi_2(x))*(C_{cos}-i*C_{sin})]\\[8pt]
  &= C + cos\varphi_2(x)*C_{cos}+sin\varphi_2(x)*C_{sin}
\end{align*}.

\section*{Acknowledgments}
This work was supported by the European Union under the European Union’s Horizon 2020 research and innovation program (grant agreements no. 101075930 to A.A. and no. 101096017 to R.V.). The copyright holder for this is those of the author(s) only and does not necessarily reflect those of the European Union or the European Research Council (ERC). Neither the European Union nor the granting authority can be held responsible for them. This work was also supported by a Joint Collaborative Research in Computational Neuroscience (CRCNS) Agence Nationale Recherche-National Science Foundation (ANR-NSF) Grant “OsciDeep” to R.V. (ANR-19-NEUC-0004) and T.S. (IIS-1912280), and an ANR Grant ANITI (ANR-19-PI3A-004) to R.V. and T.S.

 \bibliographystyle{elsarticle-harv} 
 \bibliography{references}

\end{document}